\documentclass[12pt]{article}
\usepackage{graphicx}
\usepackage{titling}
\posttitle{\par\end{center}\vskip 0.35em}
\preauthor{\begin{center}}
\postauthor{\par\end{center}\vskip -0.6em}
\predate{}
\postdate{}
\title{\fontsize{14.4}{17.3}\selectfont Evolutionary Stability Does Not Guarantee Learning Accessibility: A Multi-Agent Reinforcement Learning Perspective on Cooperation Emergence}
\author{\normalsize Yijie Wang\\[-2pt]
\small Liupanshui Normal University\\
Liupanshui, Guizhou, China\\
\texttt{wangyj@lpssy.edu.cn}}
\date{}
\begin{document}
\maketitle
\begin{abstract}
Cooperation emergence is a central problem in multi-agent systems because decentralized agents must coordinate while adapting to the changing behavior of others. Evolutionary game theory identifies strategically stable outcomes, but stability under a population adjustment dynamic need not imply that finite-sample learning agents can reach the same outcome through local reward feedback.

We study this distinction in a transparent three-agent governance-motivated game involving a government, a platform firm, and users. We derive replicator dynamics for the fixed stage-game incentives, evaluate the cooperative evolutionary basin on a symmetric initial-condition grid, and compare it with learning-basin estimates for three decentralized value-based learners. The learning analysis uses independent Q-learning with $\varepsilon$-greedy action selection, scaled Boltzmann exploration, and SA--EA BQL under the same payoff environment and outcome criterion.

The evolutionary basin has volume $V_E=1.00$ on the sampled grid. The empirical learning basin is $0.88$ for $\varepsilon$-IQL and $0.00$ for both scaled Boltzmann and SA--EA BQL. Diagnostic traces show that broader action diversity and nonzero value separation can coexist with failure to sustain the cooperative joint action in this fixed configuration.

These results indicate that evolutionary stability and learning accessibility are distinct properties of a coupled game--learning system. The shared-bike setting is a motivating application; the broader contribution is a framework for comparing population-level stability with the finite-sample accessibility of cooperation under specified multi-agent learning dynamics.
\end{abstract}
\section{Introduction}

\subsection{Motivation}

Cooperation emergence is a central problem in multi-agent systems. Autonomous decision makers increasingly operate in settings in which their returns depend on the simultaneous choices of other agents, including distributed services, robot teams, platform ecosystems, and public--private governance arrangements. In such settings, specifying a desirable collective outcome is only the first step. A system must also support the decentralized process through which agents discover and sustain compatible behavior. Multi-agent systems research has long treated this challenge as a combination of strategic interaction, information limitations, and adaptation \cite{ShohamLeytonBrown2008,StoneVeloso2000}. Multi-agent reinforcement learning (MARL) makes the adaptation problem explicit: every learner changes its behavior from reward feedback while the effective environment changes because other learners are also adapting \cite{Busoniu2008,HernandezLeal2019}.

The difficulty is especially clear in coordination problems. An agent may have an incentive to select a cooperative action only when its counterparts make compatible choices, yet early observations may be dominated by uncoordinated joint actions. Decentralized learners must therefore explore, evaluate rewards that depend on others, and form action preferences from histories that can differ substantially across runs. The shared-bike governance setting studied here provides a concrete motivating example. Government regulation, firm compliance, and user responsiveness are mutually dependent, but the substantive question is not limited to this application. It is whether a cooperative outcome that is strategically attractive can be reached by agents that start with limited information and learn only from local feedback.

This distinction matters for the design and assessment of artificial multi-agent systems. A static incentive analysis may suggest that an outcome is viable, whereas a deployed learning process may still fail to establish the joint behavior required to realize it. Conversely, a favorable learning run does not establish that a cooperative outcome is stable under changes in the population state. A useful account of cooperation emergence therefore needs to distinguish the strategic landscape from the dynamics used to navigate that landscape. This paper makes that distinction observable by evaluating population-level stability and finite-sample learning accessibility in the same fixed payoff environment.

Decentralization makes this separation practically important. A centralized controller can condition a joint decision on global information and directly enforce a coordinated policy. Independent agents instead receive local rewards, possess their own action values, and may observe only the realized consequences of a joint action. In a common-payoff environment, this does not remove the coordination problem: an individually sampled action can be reasonable under a learner's current value estimate while being incompatible with the action needed by the group. The difficulty is amplified when a cooperative reward requires several agents to select their complementary actions within the same period. Before those compatible actions have occurred repeatedly, learners may receive evidence that is sparse, variable, or dominated by the behavior induced by earlier exploration. The problem is therefore one of convention formation, not only action optimization. A system designer needs to know both whether the incentives favor cooperation and whether the specified adaptation process can form a convention that realizes those incentives.

This framing also avoids two unhelpful simplifications. First, it does not equate a cooperative equilibrium with a prediction that decentralized agents will coordinate in finite time. Second, it does not treat a realized learning outcome as a definitive description of the strategic environment. The two perspectives answer different questions and use different state variables. Evolutionary analysis tracks the distribution of strategies in a population; independent reinforcement learning tracks each agent's evolving estimates of action value. Comparing them in a common payoff environment creates a controlled way to identify where their conclusions coincide and where they diverge. Such a comparison is particularly useful for AI systems that are intended to operate without persistent central coordination, because implementation depends on the actual learning path rather than on equilibrium existence alone.

\subsection{Research gap}

Evolutionary game theory offers a principled language for analyzing strategic stability. Evolutionarily stable strategies and related stability concepts describe whether a strategy can resist invasion under specified strategic conditions \cite{MaynardSmith1982,Weibull1995}. Replicator dynamics then supplies a population-level adjustment process in which strategy shares change according to payoff differences \cite{TaylorJonker1978,HofbauerSigmund1998}. Population-game formulations further clarify how equilibrium and dynamic behavior depend on the payoff structure and the chosen revision protocol \cite{Sandholm2010}. These tools are valuable for identifying cooperative regions of a game, but they do not by themselves specify how a finite collection of independently learning agents will acquire behavior from realized rewards.

The theory of learning in games makes the adjustment process central rather than incidental. Equilibrium selection can depend on experience, perturbations, beliefs, and the particular learning rule through which agents revise behavior \cite{FudenbergLevine1998,Young1993}. In reinforcement-learning settings, an agent evaluates actions from finite histories while other agents alter the reward contingencies it encounters. This creates a distinction between a state that is attractive under a smooth population dynamic and a state that is reachable by a specified decentralized learning process. The distinction is not merely terminological. Population shares, individual action values, information flows, and sources of stochasticity differ across the two descriptions.

The comparison should consequently be made at matched levels of specificity. An evolutionary conclusion is conditional on the payoff model, the population state, and the revision dynamic used to define stability. A learning conclusion is conditional on the learner, action-selection rule, initialization, random seed, finite horizon, and criterion used to classify an outcome. Neither conditional statement subsumes the other. The relevant question is not whether one framework is more fundamental, but whether the accessible outcomes of a stated learning process correspond to the cooperative region indicated by a stated evolutionary process. This distinction has received less attention than equilibrium characterization or algorithm benchmarking, even though it is essential when an AI system must form cooperation through decentralized adaptation.

MARL provides a direct framework for examining this gap. Independent learners update their own value estimates, commonly treating the evolving behavior of other agents as part of the environment. Such decentralization is useful for studying cooperation, but it also creates non-stationarity and coordination challenges \cite{Busoniu2008,Matignon2012}. Work on learning in cooperative multi-agent systems shows that exploration, game structure, and observability can all affect which conventions are learned \cite{ClausBoutilier1998}. Yet evolutionary stability is often used as an intuitive proxy for whether cooperation should emerge, even though the proxy does not specify the learning rule, finite horizon, initialization, or outcome criterion.

The gap addressed here is therefore precise: does evolutionary stability imply learning accessibility when cooperation must emerge through finite-sample MARL? We use \emph{learning accessibility} to mean the empirical ability of a specified learning process to reach the pre-defined cooperative outcome from a defined initial-condition grid under a fixed experimental configuration. This definition is deliberately narrower than a universal claim about learnability. It allows strategic stability and learning accessibility to be compared without treating one as a substitute for the other.

\subsection{Research framework and contributions}

\begin{figure}[htbp]
\centering
\includegraphics[width=\linewidth]{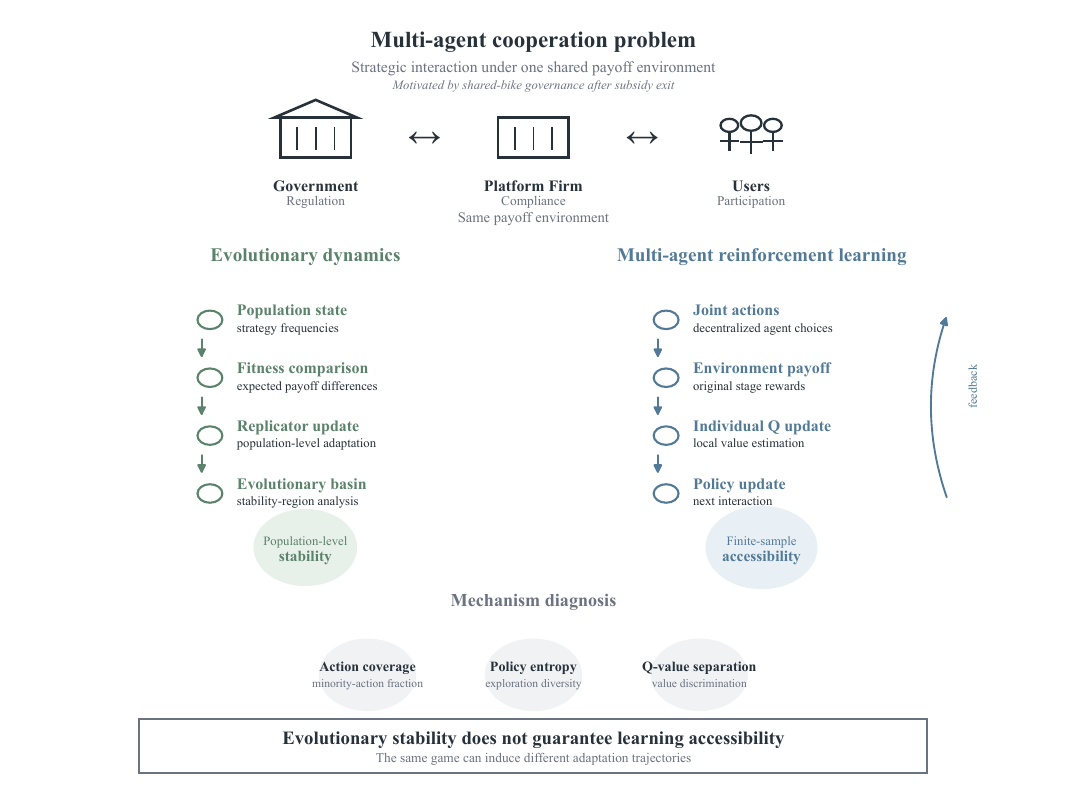}
\caption{Conceptual framework of evolutionary stability and learning accessibility. The same multi-agent payoff environment is examined through two distinct dynamics. The evolutionary branch updates population strategy frequencies through fitness comparison and replicator adjustment. The MARL branch maps joint actions to realized payoffs, individual Q-value updates, policy updates, and the next interaction.}
\label{fig:conceptual_framework}
\end{figure}

We study the question in a transparent three-agent game motivated by post-subsidy shared-bike scheduling. Government, platform firm, and user agents each choose one of two actions, and their original stage payoffs connect regulation, compliance, and responsiveness. First, we derive replicator dynamics from the expected payoff differences and evaluate the cooperative basin over a frozen symmetric grid. Second, we keep the stage game unchanged and evaluate three decentralized value-based learning dynamics: $\varepsilon$-greedy independent Q-learning (IQL), scaled Boltzmann exploration, and SA--EA BQL. Third, we compare their empirical learning basins with the evolutionary basin and use action coverage, policy entropy, and Q-value separation as diagnostics of the observed divergence. Figure~\ref{fig:conceptual_framework} summarizes this comparison.

The study makes three contributions. First, it develops an analytical framework that connects evolutionary stability with MARL accessibility while holding strategic incentives fixed. The framework treats the shared-bike setting as a motivating application rather than as evidence that its numerical outcomes generalize to every governance environment. Second, it quantifies the difference between a sampled evolutionary basin and empirical learning basins under a common initial-condition grid and outcome definition. This separates a property of the replicator system from a property of the specified learning dynamics. Third, it analyzes why exploration diversity alone does not guarantee cooperative emergence. The diagnostics show that continued access to alternative actions can coexist with failure to sustain the compatible joint action required for cooperation in the frozen configuration. Together, these contributions provide a bounded way to study how population-level stability, decentralized learning, and coordination interact.

This work is not intended as an algorithm benchmark study; instead, it investigates the relationship between equilibrium stability and accessibility under different learning dynamics.

\section{Related Work}

\subsection{Evolutionary game theory}

Evolutionary game theory links strategic interaction to dynamic selection. The concept of an evolutionarily stable strategy formalizes resistance to strategically relevant deviations \cite{MaynardSmith1982,Weibull1995}. Taylor and Jonker established the connection between evolutionary stability and game dynamics through the replicator equation \cite{TaylorJonker1978}, while Hofbauer and Sigmund developed a broad mathematical treatment of evolutionary games and population dynamics \cite{HofbauerSigmund1998}. Subsequent work has extended evolutionary analysis beyond simple normal-form settings \cite{Cressman2003}, and population-game theory has made explicit the relationship between payoff functions, revision protocols, and aggregate behavior \cite{Sandholm2010}.

This literature supplies the stability lens used in the present paper. Our contribution is not a new equilibrium concept or a modification of replicator dynamics. Instead, we use a fixed three-agent payoff model to ask whether the cooperative region identified under population adjustment is also accessible under a distinct, finite-sample learning dynamic.

\subsection{Learning in games}

Learning-in-games research studies how adaptive behavior selects, approaches, or fails to approach strategically meaningful outcomes. Fudenberg and Levine provide a foundational account of learning rules and equilibrium reasoning \cite{FudenbergLevine1998}, and Young shows how stochastic adaptation can shape the emergence of conventions \cite{Young1993}. In reinforcement-learning formulations, the relevant state representation and interaction structure matter: Markov games provide one general framework for multi-agent learning \cite{Littman1994}, while Nash Q-learning analyzes value-based learning in general-sum stochastic games \cite{HuWellman2003}. Studies of individual Q-learning in normal-form games and of independent learning in multi-agent settings further emphasize that convergence properties depend on the learning process rather than on payoffs alone \cite{LeslieCollins2005,Matignon2012}.

The present study is related to this tradition because it treats the path to a cooperative outcome as an object of analysis. It differs in its comparison target: the empirical learning basin of a specified MARL procedure is evaluated alongside a basin generated by replicator dynamics under the same strategic incentives.

\subsection{Multi-agent reinforcement learning}

MARL addresses decentralized decision making when multiple adaptive agents act in a shared environment. Early accounts distinguish independent from cooperative learning and identify the interaction between local updates and coordination \cite{Tan1993,ClausBoutilier1998}. Surveys organize the field around competing, cooperative, and mixed settings, as well as the non-stationarity induced by simultaneously learning agents \cite{Busoniu2008,HernandezLeal2019,TuylsWeiss2012}. Modern methods often use centralized information during training or structured value decomposition to improve coordination, including multi-agent actor--critic methods \cite{Lowe2017}, counterfactual policy gradients \cite{Foerster2018}, and monotonic value-factorization approaches \cite{Rashid2018}. Broader theoretical overviews likewise distinguish the information structure, learning objective, and solution concept when characterizing MARL algorithms \cite{ZhangYangBasar2021,OliehoekAmato2016}.

Recent MARL research has also studied cooperative exploration, emergent roles, learned communication, and evaluation settings that make coordination challenges visible \cite{Liu2021,Son2019,Samvelyan2019,Wang2020,Foerster2016,JiangLu2018,Leibo2017}. These directions underscore that coordination can depend on exploration, representation, communication, and population structure. They are complementary to the present controlled comparison, which does not introduce any of these mechanisms but instead asks how a fixed payoff environment behaves under specified decentralized learning dynamics.

These studies establish that coordination is sensitive to learning architecture and information design. Our paper does not introduce a new MARL algorithm or benchmark centralized methods against independent learners. It instead uses transparent decentralized value-based rules as diagnostic dynamics, asking whether the cooperative conclusion supplied by evolutionary analysis transfers to finite-sample learning accessibility. This focus also connects to the broader study of cooperation, in which repeated interaction and population processes can support cooperative conventions under conditions that must be stated explicitly \cite{Axelrod1984,Nowak2006}.
\section{Evolutionary Game Model}
\label{sec:evolutionary-model}

\subsection{Three-party governance game}

We model post-subsidy shared-bike scheduling as a repeated interaction among a government, a platform firm, and users. Each player has a binary action. The government chooses active regulation ($g=1$) or weak regulation ($g=0$); the firm chooses compliant scheduling cooperation ($f=1$) or non-compliance ($f=0$); and users choose responsive participation ($u=1$) or non-response ($u=0$). Let $x$, $y$, and $z$ denote the population shares choosing $g=1$, $f=1$, and $u=1$, respectively. This representation uses the evolutionary-game perspective on strategic stability while retaining a tractable multi-population payoff structure \cite{MaynardSmith1982,HofbauerSigmund1998}.

The verified stage-game payoff matrix is given in Table~\ref{tab:payoff-matrix}. Regulation incurs cost $C_g$ and yields the government a subsidy-efficiency return $\theta S$. A compliant firm receives the residual subsidy benefit $\delta B$ but bears compliance cost $c$; a non-compliant firm obtains the private gain $E$ and is exposed to a sanction $F$ detected with probability $\lambda_s$ under active regulation and $\lambda_w$ under weak regulation. User responsiveness produces baseline net utility $a-c_u$ and may receive firm and government benefits $q$ and $r$. The symbols $S$, $L$, $\beta$, $\pi_e$, and $\alpha$ retain their verified meanings in the implementation: public scheduling benefit, loss from non-compliance, government responsiveness benefit, the firm's operating payoff, and the firm--user coordination return.

\subsection{Definition: Learning Accessibility}

Let $M$ denote a specified learning mechanism and let $C_i=1$ if independent run or initial state $i$ reaches the pre-defined cooperative outcome under the fixed evaluation criterion, and $C_i=0$ otherwise. For $N$ evaluated runs or initial states, we define the empirical learning-accessibility measure as
\begin{equation}
A_L(M)=\frac{1}{N}\sum_{i=1}^{N}\mathrm{I}(C_i=1).
\label{eq:learning-accessibility}
\end{equation}
This quantity records the observed fraction of successful learning trajectories for the stated mechanism, initialization design, finite horizon, and stochastic-update procedure. It is an empirical accessibility measure, not a universal metric of learnability.

The evolutionary basin instead records whether a stable state attracts the population dynamics from the specified initial states. The two quantities therefore refer to different adaptation processes. Evolutionary dynamics updates strategy frequencies through population-level payoff comparison and replicator adjustment. Learning dynamics updates individual value estimates and policies from finite samples of joint interaction. Consequently, an evolutionary basin need not equal the learning basin measured by $A_L(M)$, even when both analyses use the same payoff environment.

\begin{table}[htbp]
\centering
\caption{Verified stage-game payoffs $(\Pi_G,\Pi_F,\Pi_U)$.}
\label{tab:payoff-matrix}
\begin{tabular}{c c c l}
\hline
$g$ & $f$ & $u$ & $(\Pi_G,\Pi_F,\Pi_U)$\\
\hline
0&0&0&$(-L+\lambda_wF,\ \pi_e+E-\lambda_wF,\ 0)$\\
0&0&1&$(-L+\lambda_wF+\beta,\ \pi_e+E-\lambda_wF,\ a-c_u)$\\
0&1&0&$(S-\delta B,\ \pi_e-c+\delta B,\ 0)$\\
0&1&1&$(S-\delta B+\beta,\ \pi_e-c+\delta B+\alpha,\ a-c_u+q)$\\
1&0&0&$(-C_g+\theta S-L+\lambda_sF,\ \pi_e+E-\lambda_sF,\ 0)$\\
1&0&1&$(-C_g+\theta S-L+\lambda_sF+\beta,\ \pi_e+E-\lambda_sF,\ a-c_u+r)$\\
1&1&0&$(-C_g+\theta S+S-\delta B,\ \pi_e-c+\delta B,\ 0)$\\
1&1&1&$(-C_g+\theta S+S-\delta B+\beta,\ \pi_e-c+\delta B+\alpha,\ a-c_u+q+r)$\\
\hline
\end{tabular}
\end{table}

\subsection{Replicator dynamics}

For each population, strategy shares evolve in proportion to the payoff advantage of the active strategy. This replicator formulation follows the standard payoff-difference interpretation of evolutionary adjustment \cite{TaylorJonker1978,Sandholm2010}. Taking expectations over the other two populations yields
\begin{equation}
\dot{x}=x(1-x)\left[-C_g+\theta S+(\lambda_s-\lambda_w)F(1-y)\right].
\label{eq:rep-gov}
\end{equation}
\begin{equation}
\dot{y}=y(1-y)\left[-c+\delta B+\alpha z-E+
\left\{\lambda_w+(\lambda_s-\lambda_w)x\right\}F\right].
\label{eq:rep-firm}
\end{equation}
\begin{equation}
\dot{z}=z(1-z)\left(a-c_u+qy+rx\right).
\label{eq:rep-user}
\end{equation}
These equations preserve the binary-action boundaries and separate the incentives associated with regulation, firm compliance, and user responsiveness. They are evaluated with the verified parameterization and numerical integration routine used throughout the project.

\subsection{Stability and basin evaluation}

Local stability characterizes whether a perturbation near a candidate equilibrium decays under equations~\ref{eq:rep-gov}--\ref{eq:rep-user}; it does not establish that a learning process will reach that equilibrium. We therefore complement the local analysis with a numerical basin calculation. Starting from the frozen symmetric grid
\[
(x_0,y_0,z_0)=(p,p,p), \qquad
p\in\{0.1,0.2,\ldots,0.9\},
\]
we integrate the replicator system under the incentive-on regime ($\alpha=80$) and classify a trajectory as cooperative when it converges to the cooperative corner according to the project criterion. All nine initial conditions converge to that corner, giving the evolutionary basin estimate
\[
V_E=\frac{9}{9}=1.00.
\]
This is a numerical statement for the sampled symmetric grid, rather than a claim about every point in the continuous state space.
\section{Multi-Agent Reinforcement Learning Framework}
\label{sec:marl-framework}

\subsection{Multi-agent environment}

The learning environment implements the same verified three-party stage game as Table~\ref{tab:payoff-matrix}. At each round, the government, firm, and user independently select one of their two actions, producing a joint action $(g,f,u)$ and the corresponding vector of original stage rewards $(\Pi_G,\Pi_F,\Pi_U)$. This decentralized formulation follows the multi-agent view in which other adaptive agents contribute to the learning environment \cite{Littman1994,ZhangYangBasar2021}. No reward shaping, auxiliary coordination bonus, or altered payoff is introduced for the learning experiments.

The static basin experiments use a fixed favourable subsidy state ($\delta=1$) and repeatedly sample the same stage-game incentives. The environment interface also records subsidy and behavioural summaries for dynamic diagnostic runs, but the Phase~G basin comparison is deliberately a fixed-state accessibility test. The initial-preference grid is imposed only at the first decision; subsequent behaviour is determined by each learner's own updates and exploration rule.

\subsection{Independent Q-learning}

Each player $i\in\{G,F,U\}$ maintains action values $Q_i(a_i)$ and treats the actions of the other two players as part of the environment. This value-based update is a repeated-stage specialization of the Q-learning tradition \cite{WatkinsDayan1992,SuttonBarto2018}. After observing its original stage reward $r_{i,t}$, it updates the selected action according to
\begin{equation}
Q_{i,t+1}(a_{i,t}) =
Q_{i,t}(a_{i,t})+
\eta_t\left[r_{i,t}-Q_{i,t}(a_{i,t})\right],
\label{eq:iql-update}
\end{equation}
while leaving the unselected action unchanged. This is a repeated-stage formulation: no continuation-value term is used in equation~\ref{eq:iql-update}. Learning rates, run lengths, random seeds, and other implementation settings are those recorded in the frozen experiment configuration rather than additional manuscript assumptions.

The $\varepsilon$-greedy independent Q-learning (IQL) benchmark chooses a currently highest-valued action except for its configured exploration probability. Independent learners are a deliberately transparent MARL baseline, but their local reward updates can encounter coordination and non-stationarity challenges \cite{Busoniu2008,Matignon2012}. Its role is descriptive: it provides a reference learning dynamic for comparing basin accessibility, not an optimized policy.

\subsection{Scale-aware Boltzmann exploration}

The scale-aware Boltzmann learner samples an action using a softmax transformation of action values after normalizing for the payoff scale,
\begin{equation}
\Pr_i(a\mid s)=
\frac{\exp\!\left(Q_i(s,a)/[\tau_i\,d_i]\right)}
{\sum_{a'\in\{0,1\}}\exp\!\left(Q_i(s,a')/[\tau_i\,d_i]\right)},
\label{eq:scaled-softmax}
\end{equation}
where $\tau_i$ is the configured temperature and $d_i$ is the payoff-scale normalization used by the implementation. The normalization makes the softmax response interpretable when the three players' payoff ranges differ. It is evaluated as a diagnostic comparison with IQL, not presented as a superior algorithm.

\subsection{Adaptive exploration variant (SA--EA BQL)}

The SA--EA BQL adaptive exploration variant combines scale-aware action selection with an exploration adjustment based on the learner's action-value separation. In generic form, its temperature is allowed to depend on a Q-value gap $\Delta Q_i$,
\begin{equation}
\tau_i=\tau_{0,i}+k_i\Delta Q_i,
\label{eq:adaptive-temperature}
\end{equation}
using the fixed implementation configuration. This diagnostic comparison mechanism is included to examine whether adaptive exploration changes learning accessibility. It is not a claim that adaptive exploration guarantees cooperation, nor are its diagnostic outcomes interpreted as causal evidence beyond the frozen comparison.
\section{Experimental Protocol}

The learning-basin comparison uses a fixed protocol so that differences across learning mechanisms are not confounded with changes in the environment or evaluation design. Table~\ref{tab:experimental-protocol} records the configuration used throughout the reported comparison.

\begin{table}[htbp]
\centering
\caption{Fixed experimental protocol for the learning-basin comparison.}
\label{tab:experimental-protocol}
\begin{tabular}{l l}
\hline
Item & Configuration\\
\hline
Agents & 3 (government, platform firm, users)\\
Action space & Binary action for each agent\\
Training horizon & 20,000 episodes per run\\
Random seeds & 50\\
Initial conditions & Symmetric initial-preference grid\\
Algorithms & $\varepsilon$-IQL; scaled Boltzmann; SA--EA BQL\\
\hline
\end{tabular}
\end{table}

All mechanisms interact with the same fixed stage-game payoffs and are evaluated using the same pre-specified cooperative-outcome criterion. The protocol is reported for reproducibility; it does not represent an additional parameter search or an algorithm-tuning exercise.
\section{Results}

\subsection{Evolutionary stability on the sampled grid}

Under the incentive-on specification ($\alpha=80$), numerical integration of the replicator system converges to the cooperative corner from each of the nine frozen symmetric starting points. The resulting sampled evolutionary basin is therefore $V_E=1.00$ (9/9). This result establishes evolutionary accessibility for the specified grid and parameterization; it is not a global basin proof.

\begin{figure}[htbp]
\centering
\includegraphics[width=0.9\linewidth]{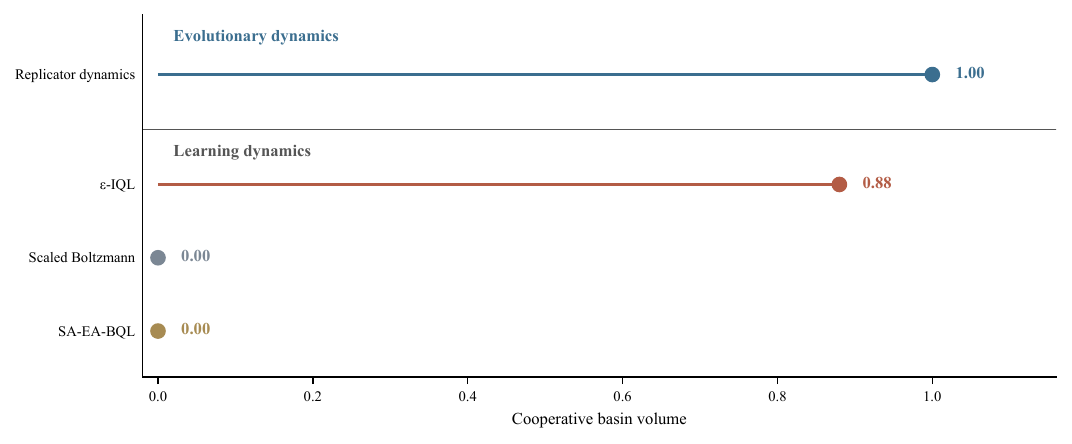}
\caption{Comparison between evolutionary stability and learning accessibility. Although the replicator dynamics reaches the cooperative basin for all evaluated initial conditions, finite-sample reinforcement learning exhibits learning-mechanism-dependent accessibility.}
\label{fig:evolutionary-learning-accessibility}
\end{figure}

\begin{figure}[htbp]
\centering
\includegraphics[width=\linewidth]{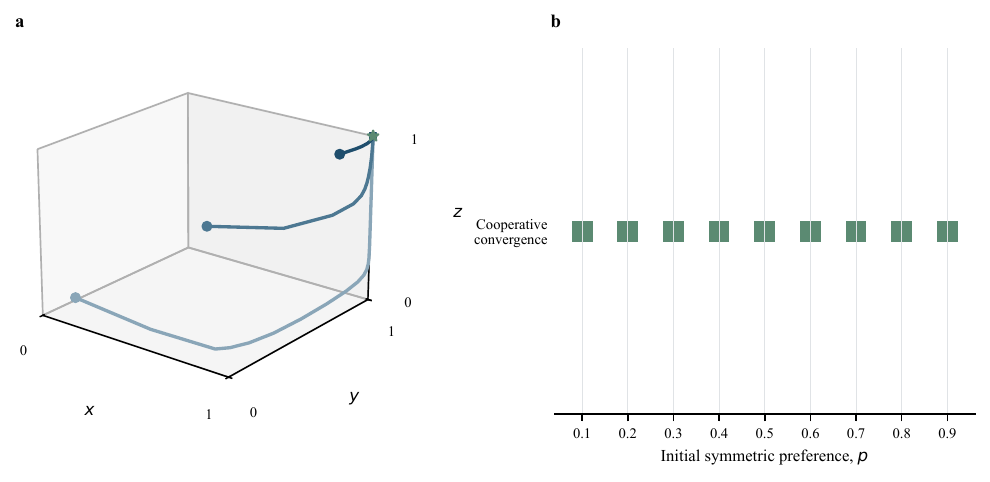}
\caption{Replicator stability on the frozen symmetric initial-condition grid. (a) Three verified trajectories project toward the cooperative corner $(1,1,1)$. (b) Each evaluated value of $p$ converges cooperatively; this discrete classification does not infer a continuous basin.}
\label{fig:replicator-stability}
\end{figure}

\subsection{Learning accessibility}

The corresponding reinforcement-learning experiment contains 1,350 runs: nine initial-preference settings, 50 random seeds per setting, and three learning rules. A run is classified using the frozen cooperative-outcome criterion. Epsilon-greedy IQL reaches cooperation in 396 of 450 runs, yielding a learning-basin estimate of $0.880$ with bootstrap 95\% confidence interval $[0.849,0.909]$. The estimated tipping preference is $p=0.1$ under the recorded rule.

\begin{figure}[htbp]
\centering
\includegraphics[width=\linewidth]{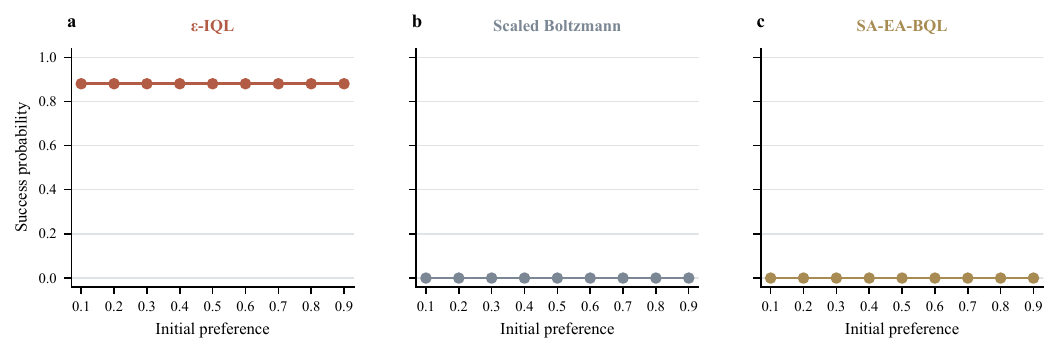}
\caption{Learning-basin curves for the three frozen reinforcement-learning comparisons. Points are observed success probabilities at the nine evaluated symmetric initial preferences; no interpolation is used.}
\label{fig:learning-basin}
\end{figure}

\begin{figure}[htbp]
\centering
\includegraphics[width=0.9\linewidth]{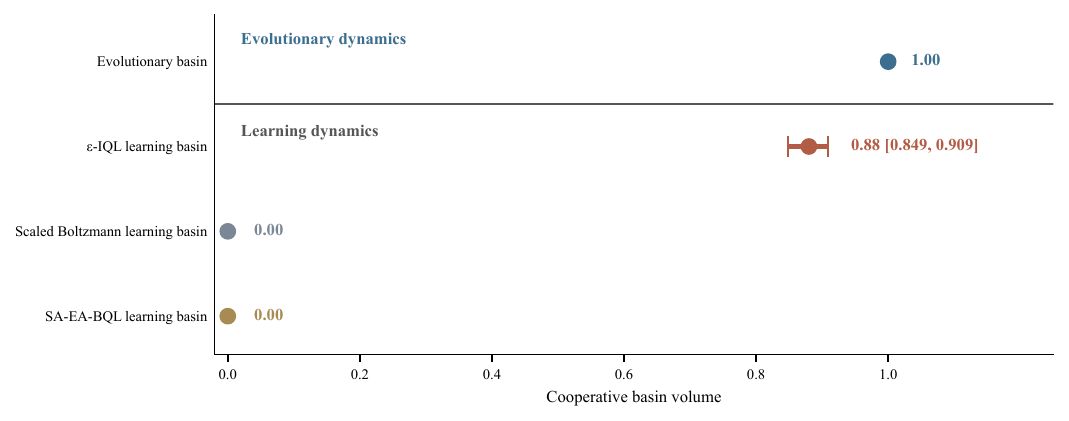}
\caption{Theory--learning gap in cooperative basin volume. The interval for $\varepsilon$-IQL is the frozen bootstrap 95\% confidence interval; the two zero estimates are exact outcomes in their respective 450-run grids.}
\label{fig:basin-gap}
\end{figure}

By contrast, neither scale-aware Boltzmann nor SA--EA BQL reaches the cooperative outcome in any of its 450 runs: both have an estimated learning basin of $0.000$. Thus, within this fixed design, evolutionary convergence does not imply that every decentralized learning rule accesses the same cooperative outcome. Figure~\ref{fig:evolutionary-learning-accessibility} distinguishes the evolutionary result from the learning-mechanism-dependent estimates; Figures~\ref{fig:learning-basin} and~\ref{fig:basin-gap} give the disaggregated learning curves and the theory--learning gap.

\subsection{Exploration mechanism diagnosis}

Phase~H records action trajectories, action coverage, policy entropy, and Q-value separation for 50 seeds per learning rule over 20,000 rounds. The diagnostic traces show that the two Boltzmann-based learners retain substantial action diversity rather than collapsing immediately into a single deterministic policy. Their Q-value gaps are also nonzero. These observations rule out a simple account in which the zero-basin results arise solely from premature deterministic-policy collapse.

A more cautious interpretation is that persistent stochastic action selection can impede convention formation in this coordination environment: learners may continue to visit both actions without jointly sustaining the mutually cooperative action profile. This is a mechanism diagnosis tied to the observed trajectories, not evidence that exploration is intrinsically harmful or that one algorithm is universally preferable.

\begin{figure}[htbp]
\centering
\includegraphics[width=\linewidth]{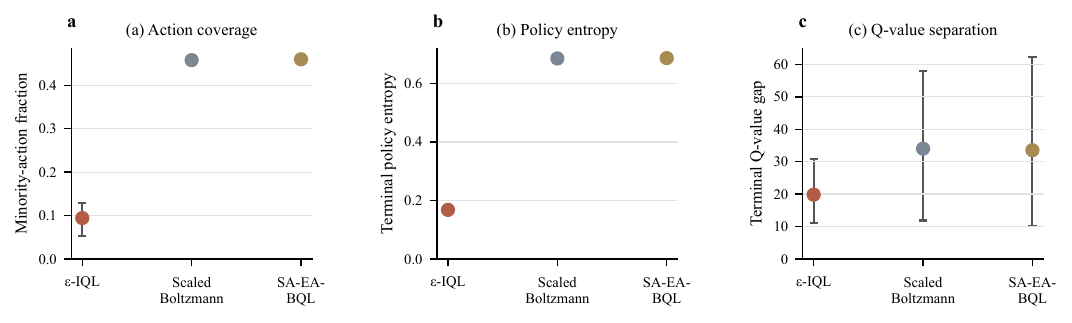}
\caption{Exploration diversity alone does not determine cooperative emergence. Points summarize the three players within each seed; error bars show empirical 2.5th--97.5th percentiles across 50 seeds. Action coverage is measured over the recorded run, whereas entropy and Q-value separation use the terminal logged episode.}
\label{fig:exploration-mechanism}
\end{figure}

\section{Discussion}

\subsection{Evolutionary stability versus learning accessibility}

The central finding is that cooperative stability under the replicator system and cooperative accessibility under finite-sample reinforcement learning are related but distinct properties. On the frozen symmetric grid, the evolutionary analysis reaches the cooperative basin for all evaluated initial conditions. The learning comparison, however, yields method-dependent basin estimates under the same underlying strategic environment. This contrast does not imply that either framework is defective. Instead, it reflects that they instantiate different dynamical processes. Replicator dynamics updates population shares according to payoff advantages that are evaluated against the current population state. The learning agents update action values from realized, stochastic stage rewards while their counterparts are simultaneously adapting. The state variables, information available at each update, and effective averaging of payoff feedback are consequently different. A cooperative fixed point that attracts nearby population states can therefore remain difficult for decentralized learners to reach from finite histories.

This distinction clarifies what is and is not established by the present results. The evolutionary basin estimate of $1.00$ is a numerical result for nine sampled symmetric initial conditions under the incentive-on parameterization; it is not a claim that every continuous initial state converges cooperatively. Likewise, the learning-basin estimates describe the observed outcomes of the specified finite grids, seeds, learning rates, and exploration rules. Within those boundaries, strategic incentives sufficient for evolutionary convergence are not sufficient to ensure that every learning mechanism forms the corresponding cooperative convention. Figure~\ref{fig:evolutionary-learning-accessibility} makes this separation visible, while Figures~\ref{fig:replicator-stability}--\ref{fig:basin-gap} distinguish the population calculation, learning curves, and IQL uncertainty. Cooperation should be assessed not only by its stability in a game-theoretic model, but also by whether a stated adaptation process can access it.

One interpretation is that learning accessibility depends on a sequence of compatible events rather than solely on the eventual payoff ordering of actions. A learner must sample an action, experience an informative reward, update its value estimate in a favorable direction, and encounter counterparts whose actions make that experience repeatable. In a coordination environment, each of these steps is coupled to the other agents' changing policies. A temporary cooperative joint action may not be observed often enough, or consistently enough, to become a stable convention in all learning processes. Conversely, an evolutionary equation effectively summarizes payoff advantages at the population level and does not represent the finite sequence of joint actions through which individual values are acquired. The results therefore suggest that the path to cooperation is itself part of the scientific object. Future theoretical work could seek conditions under which properties of a payoff game and properties of a particular learning update are aligned, rather than assuming that equilibrium attraction transfers directly from one dynamic to the other.

This distinction is consistent with the broader view that stability and learning are properties of specified dynamics, rather than of a payoff table in isolation \cite{HofbauerSigmund1998,FudenbergLevine1998}.

\subsection{Exploration and cooperative convention formation}

The mechanism diagnostics refine a simple explanation of the zero-basin outcomes. It would be tempting to attribute non-cooperation by the Boltzmann-based learners to insufficient exploration or to premature collapse into one deterministic action. The recorded Phase~H traces do not support that simple account. As summarized in Figure~\ref{fig:exploration-mechanism}, the Boltzmann-based learners retain broader action diversity than the $\varepsilon$-greedy benchmark on the reported minority-action, terminal-entropy, and Q-value-gap diagnostics. Their nonzero Q-value separation also indicates that the learners are not merely indifferent between their actions. Yet these properties coexist with an inability to sustain the cooperative outcome under the frozen basin criterion.

This pattern suggests a more specific, and more cautious, interpretation. Exploration diversity can preserve access to alternative actions, but access is not the same as coordinated commitment. In the present game, a cooperative convention requires compatible actions by all three agents. Continued stochastic action selection may repeatedly interrupt the joint profile that would otherwise reinforce cooperative values, particularly when each learner treats the decisions of the others as part of a changing environment. The diagnostic evidence is consistent with persistent exploration impeding convention formation in this particular design, but it does not identify a unique causal mechanism. Other features of the learning process, including relative payoff scales, update trajectories, early joint-action histories, and the form of value normalization, may also contribute. The comparison should therefore not be interpreted as evidence that Boltzmann exploration is generally unfavorable, or that lower action diversity is generally desirable.

The distinction between diversity and coordination has practical consequences for how exploration is evaluated in cooperative MARL. A high entropy or minority-action fraction can be useful evidence that agents have not become behaviorally trapped, but these metrics do not measure whether agents are learning compatible policies. Similarly, a nonzero value gap documents action differentiation without indicating whether the differentiated preferences support the same joint convention across agents. When cooperation requires synchronized behavior, a diagnostic suite should therefore combine individual-level indicators with an outcome-level measure of sustained joint coordination. The learning-basin criterion used here provides one such outcome-level measure for a fixed environment. More detailed analyses could track the temporal formation and disruption of candidate conventions, distinguish transient coordination from persistent coordination, and examine which joint-action histories precede successful runs. Such analyses would help separate exploration that is informative for coordination from exploration that remains behaviorally diverse without building a shared convention.

This coordination challenge is characteristic of multi-agent learning, in which each learner faces an environment altered by the adaptation of other learners \cite{Busoniu2008}.

\subsection{Implications for multi-agent systems}

The results suggest several implications for the design and assessment of cooperative multi-agent systems. First, equilibrium analysis alone is an incomplete basis for evaluating whether a proposed incentive structure will yield cooperation in a learning population. It remains valuable for identifying strategically stable states and for understanding how payoff components shape those states. However, a stable solution does not by itself reveal the learning trajectory by which bounded agents will approach it. Evaluations of cooperative systems may therefore benefit from reporting both a strategic analysis and a learning-accessibility analysis, with the latter specifying the initialization, update rule, exploration process, and finite evaluation horizon. This paired perspective is relevant wherever system designers rely on decentralized adaptation rather than direct coordination.

Second, mechanism design and learning design should be considered jointly. A policy instrument, reward structure, or sanction may create a cooperative equilibrium while leaving the learning path fragile or inaccessible for some adaptive agents. Conversely, a learning rule can alter which strategically available outcomes are observed in finite time without changing the underlying game. The present framework does not prescribe a universal intervention, but it provides a way to diagnose this mismatch. A mechanism can be assessed by the basin it creates under evolutionary adjustment and by the extent to which specified learners access that basin. When the two assessments diverge, the discrepancy points to a design question: whether to modify incentives, information, communication, or the adaptation process itself. The answer will depend on the application and should be tested rather than inferred from the present three-agent model.

Third, learning accessibility is useful as a reporting dimension rather than as a post hoc explanation for an unsuccessful run. Finite time, finite data, and exploration are operational constraints in many multi-agent settings. Basin-based comparisons connect equilibrium predictions to observed learning outcomes through a common outcome definition while preserving their distinct dynamics. Claims about cooperative behavior should therefore state the adjustment process under which they are expected to hold.

Accordingly, strategic and learning dynamics should be reported as complementary objects of analysis when assessing decentralized multi-agent systems \cite{Sandholm2010,ShohamLeytonBrown2008}.

\subsection{Relation to learning in games}

The comparison can also be positioned within learning in games. Replicator dynamics describes population-level adaptation: strategy shares change according to expected payoff differences at the current population state. MARL, by contrast, describes individual value-based adaptation: each agent updates action values from realized rewards while the behavior of other agents changes the data-generating process. These dynamics can agree in some settings, but their agreement is not guaranteed by the existence of a stable equilibrium.

This difference helps explain why a cooperative equilibrium may be stable without automatically becoming a reachable learning outcome. A population equation aggregates payoff advantages, whereas a finite learner must experience compatible joint actions often enough to form and retain a favorable value ordering. The learning-in-games literature similarly treats the adjustment rule and the path of adaptation as part of equilibrium discovery rather than as details that can be omitted \cite{FudenbergLevine1998,Young1993,LeslieCollins2005}. In the present fixed configuration, the basin comparison makes this distinction measurable without claiming that one learning rule is generally superior to another.

\subsection{Limitations and future directions}

Several limitations delimit the scope of the present inference. The current study focuses on a controlled analytical setting: one transparent three-agent governance game with binary actions and fixed payoff incentives. Its value is that it isolates a tractable interaction among regulation, compliance, and responsiveness, but its structure does not represent every form of multi-agent cooperation. Larger populations, heterogeneous agent types, network interactions, and richer action spaces may generate different relationships between evolutionary stability and learning accessibility. The central within-design comparison remains informative for the specified game, but its transportability to other strategic structures requires additional tests.

Second, the numerical results depend on finite parameter configurations, a sampled symmetric initial-condition grid, and a fixed evaluation horizon. The evolutionary basin is therefore a grid-based numerical result, and the learning estimates are finite-sample summaries rather than exhaustive characterizations. Broader parameter sweeps, asymmetric initial conditions, and sensitivity analyses could test whether the gap persists across other payoff relationships and learning preferences.

Third, the study does not include human behavioral experiments. The agents are computational learners whose reward updates and exploration rules are specified by the model. The results consequently address dynamical accessibility in a MARL setting, not how human participants would respond to regulatory, organizational, or user incentives. Human experiments, field observations, or hybrid designs could test whether analogous coordination barriers arise when learning is shaped by social beliefs, communication, or institutional knowledge not represented here.

Finally, the comparison does not establish that any reinforcement-learning algorithm dominates across settings. The three learners are diagnostic instruments applied to a common environment, not a comprehensive benchmark suite. The two zero-basin outcomes and the positive IQL basin should not be generalized beyond the frozen configuration. Future extensions may consider continuous state spaces, deep function approximation, larger agent populations, and richer communication mechanisms. A useful next step is a theoretical characterization of learning basins under these extensions, including when information sharing, commitment devices, or alternative update structures change accessibility.
\section{Conclusion}

This study examines a simple but consequential question for cooperation emergence in multi-agent systems: does a cooperative state that is stable under evolutionary dynamics necessarily emerge when decentralized agents learn from finite, stochastic experience? We addressed this question by combining a three-agent evolutionary game with controlled multi-agent reinforcement-learning comparisons. The governance setting supplies a concrete interaction among regulation, firm compliance, and user responsiveness, but the analytical objective is broader. It is to separate a strategic property of a game from a dynamical property of a learning process. The resulting framework places evolutionary game theory and MARL in a common environment while preserving the differences between population-level strategy adjustment and individual value-based adaptation.

The evolutionary analysis identifies a cooperative state that is accessible from all nine evaluated symmetric initial conditions in the incentive-on specification, producing a sampled basin estimate of $V_E=1.00$. The reinforcement-learning results do not mirror this outcome uniformly. Under the frozen learning-basin design, $\varepsilon$-greedy IQL reaches the cooperative criterion in 396 of 450 runs, whereas the scale-aware Boltzmann and SA--EA BQL comparisons do not reach that criterion in their respective grids. These estimates should be interpreted within the stated game, initial-preference grid, seeds, and finite evaluation horizon. They do not establish a global property of any algorithm. Nevertheless, their contrast with the evolutionary calculation provides direct evidence that a stable cooperative state can be learning-mechanism-dependent in finite-sample decentralized adaptation.

The mechanism diagnosis further qualifies how this gap should be interpreted. The Boltzmann-based learners retain action diversity and nonzero Q-value separation in the recorded diagnostic runs. Their lack of cooperative convergence therefore cannot be reduced to a simple story of insufficient exploration or immediate deterministic-policy collapse. Instead, the evidence suggests that exploration diversity alone does not ensure the formation of a stable cooperative convention. In a coordination problem, learners must not only preserve alternative actions; they must also repeatedly realize compatible joint actions long enough for their individual updates to support the same collective pattern. Persistent stochasticity may interrupt this process in the present configuration, although the current evidence does not isolate a unique causal route.

The main conclusion is consequently bounded but general in its conceptual implication: stable cooperation in evolutionary dynamics does not automatically emerge from finite-sample reinforcement learning. Equilibrium analysis remains useful for identifying strategically viable cooperative states, and MARL analysis remains useful for observing how specified agents adapt to realized feedback. Neither perspective substitutes for the other. Jointly reporting an evolutionary basin and a learning basin makes the difference visible and provides a practical way to assess whether a cooperative mechanism is not only stable in principle but accessible to the learning process expected to implement it.

Several directions follow from this boundary. Future work could test larger and more heterogeneous populations, allow richer communication or commitment mechanisms, and characterize learning basins analytically rather than only through finite grids. It could also compare computational learning with human behavior in settings where beliefs, norms, and institutional information affect coordination. These extensions are needed before broad claims about governance systems or MARL algorithms can be made. Within the present design, however, the results clarify a central point: explaining cooperation requires attention both to the incentives that stabilize an outcome and to the learning dynamics that determine whether agents can reach it.

\appendix
\section{Mathematical details}

\subsection{Derivation of the replicator dynamics}

For a binary-action population with active-strategy share $w$, let $\pi_1$ and $\pi_0$ denote the expected payoffs of the active and inactive actions. The average payoff is $\bar{\pi}=w\pi_1+(1-w)\pi_0$. The standard binary replicator identity is therefore
\[
\dot{w}=w(\pi_1-\bar{\pi})=w(1-w)(\pi_1-\pi_0).
\]
Applying this identity to each population and taking expectations over the other two strategy shares yields the three equations in Section~2.

For the government, active regulation differs from weak regulation only through its direct regulatory return and the sanction applied when the firm is non-compliant. Hence,
\[
\pi_G(g=1)-\pi_G(g=0)
=-C_g+\theta S+(\lambda_s-\lambda_w)F(1-y).
\]
Substitution into the binary identity with $w=x$ gives equation~\ref{eq:rep-gov}.

For the firm, compliance replaces the private gain $E$ and avoided sanction with the compliance cost and residual subsidy benefit; the firm--user coordination return is received when users are responsive. Averaging the sanction term over the government share gives
\[
\pi_F(f=1)-\pi_F(f=0)
=-c+\delta B+\alpha z-E+
\left\{\lambda_w+(\lambda_s-\lambda_w)x\right\}F.
\]
Using $w=y$ gives equation~\ref{eq:rep-firm}.

For users, non-response has zero payoff in the specified stage game. Responsive participation has the baseline net return $a-c_u$, the firm-related return $q$ with probability $y$, and the government-related return $r$ with probability $x$. Thus,
\[
\pi_U(u=1)-\pi_U(u=0)=a-c_u+qy+rx.
\]
Using $w=z$ gives equation~\ref{eq:rep-user}. These derivations introduce no additional behavioral assumptions; they only rewrite the payoff differences already encoded in Table~\ref{tab:payoff-matrix}.
\section{Reinforcement-learning algorithm details}

\subsection{Algorithm 1: Independent Q-learning training procedure}

For each run, initialize each agent's two action values from the fixed initial-preference condition and retain the run's assigned random seed. Then repeat the following procedure for the configured number of episodes.

\begin{enumerate}
\item For each agent $i\in\{G,F,U\}$, select a binary action $a_{i,t}$ using that learner's configured action-selection rule.
\item Execute the joint action $(g,f,u)$ in the unchanged stage game and observe the original individual reward $r_{i,t}$ for every agent.
\item Update only the selected action value according to equation~\ref{eq:iql-update}; leave the unselected action value unchanged.
\item Continue until the fixed episode horizon is reached. Classify the completed run using the pre-specified learning-basin criterion.
\end{enumerate}

This procedure is independent Q-learning: each agent maintains its own values and does not observe, communicate, or centrally optimize the values of the other agents. Because the environment is repeated-stage, the update has no continuation-value term.

\subsection{Alternative action-selection rules}

The scaled Boltzmann learner uses the normalized softmax probability in equation~\ref{eq:scaled-softmax}. Its temperature and payoff-scale normalization are the fixed values used in the frozen implementation. The SA--EA BQL adaptive exploration variant uses the same scale-aware choice rule while adjusting its temperature from the action-value separation according to equation~\ref{eq:adaptive-temperature}. These alternatives are diagnostic comparison mechanisms included to examine whether adaptive exploration changes learning accessibility. Their description does not imply that either rule is generally superior or more appropriate outside the present fixed configuration.
\section{Experimental configuration and reproducibility}

Table~\ref{tab:reproducibility} records the fixed configuration used for the learning-basin comparison. It is provided to make the reported finite-sample estimates reproducible and should not be read as an additional parameter search.

\begin{table}[htbp]
\centering
\caption{Fixed configuration for the learning-basin comparison.}
\label{tab:reproducibility}
\begin{tabular}{l l}
\hline
Item & Fixed configuration\\
\hline
Number of agents & 3 (government, platform firm, users)\\
Action space & Binary action for each agent\\
Episodes per run & 20,000\\
Random seeds & 50\\
Initial-preference grid & $0.1,0.2,\ldots,0.9$ (symmetric)\\
Algorithms & $\varepsilon$-IQL; scaled Boltzmann; SA--EA BQL\\
Evaluation & Pre-specified learning-basin criterion\\
\hline
\end{tabular}
\end{table}

Each algorithm was evaluated over the same grid and seed structure in the unchanged binary-action payoff environment. The learning-basin criterion is the outcome definition used in the main text: it records whether a run reaches the specified cooperative outcome under the fixed evaluation design.

\bibliographystyle{plain}
\bibliography{references}
\end{document}